# Algorithmic Ghost in the Research Shell: Large Language Models and Academic Knowledge Creation in Management Research

**Nigel Williams**
The School of Organisations, Systems and People; Faculty of Business and Law; University of Portsmouth, UK. email: nigel.williams@port.ac.uk

**Stanislav Ivanov**
1. Professor and Vice-Rector (Research), Varna University of Management,13A Oborishte str., 9000 Varna, Bulgaria, email: stanislav.ivanov@vumk.eu, web: http://stanislavivanov.com/
2. Director, Zangador Research Institute, 9010 Varna, Bulgaria, email: info@zangador.institute

**Dimitrios Buhalis**
Business School, Bournemouth University, UK, email: dbuhalis@bournemouth.ac.uk

**Abstract:**
The paper looks at the role of large language models in academic knowledge creation based on a scoping review (2018 to January 2023) of how researchers have previously used the language model GPT to assist in the performance of academic knowledge creation tasks beyond data analysis. These tasks include writing, editing, reviewing, dataset creation and curation, which have been difficult to perform using earlier ML tools. Based on a synthesis of these papers, this study identifies pathways for a future academic research landscape that incorporates wider usage of large language models based on the current modes of adoption in published articles as a CoWriter, Research Assistant and Respondent. The paper concludes with a research and practice agenda for management knowledge creation based on the wider adoption of Large Language models. The paper's focus is on understanding the nature of the current usage of GPT to perform academic tasks. As such, it does not describe the challenges and problems of large language models. It does also not speculate about the extent to which they present machine intelligence or consciousness.

**1. Introduction**

Artificial Intelligence (AI) has been defined as the research and design of creating machines that simulate human intelligence to perform actions or intellectual tasks (Müller and Bostrom, 2016). This paper will define AI simply as "computer systems that perform tasks requiring cognition tasks autonomously". This is similar to earlier definitions (Russell, 2010).

Emerging phenomena can often be overlooked in management research as they are poorly defined with unclear conceptual concepts and limited empirical data (Yadav, 2018). Large Language models like GPT, however, have growth drivers that suggest that they are worthy of researcher attention and specifically, their impact on academic knowledge production should be identified at this early stage of adoption.

Previous academic research in business and management have identified the potential for machine learning analytics to change the nature of theorising in business and management reseach (Leavitt, Schabram, Hariharan & Barnes, 2021). Large Language models such as GPT, however, can go further to influence the nature of academic knowledge production itself in this domain. Management research is primarily based on empirical quantitative and qualitative studies done by small teams of researchers which may be difficult to replicate (Block, Fisch, Kanwal, Lorenzen, & Schulze, 2022). As a result, the domain can be influenced by tools that can simulate human created text in a manner that experimental or lab based work would not. The area therefore requires examination which is the purpose of this paper.

## 2. AI transformations

AI in the Business and management domain research has taken two main perspectives. The first is as an analytical approach: AI tools identify insights, using classification or modelling of complex dynamic data. AI analytical approaches enable the direct examination of "mixed" data, such as data collected from social media that can combine text, images and video (Al-Smadi, Jaradat, Al-Ayyoub, and Jararweh, 2017). These approaches have been used to examine the meaning of text (Martinez-Torres, and Toral, 2019), identify market segments via geographical data (Rodríguez, Semanjski, Gautama, Van de Weghe and Ochoa, 2018) and emergent visual representations from photographs (Zhang, Chen, and Li, 2019). The second stream of research examines the impact of AI on organisational activity. Research has examined the extent to which AI can be applied in service operations (Meyer, Cohen, and Nair, 2020).

Machine learning (ML) is the dominant approach to implementing artificial intelligence in computer systems (Ghahramani, 2015). Neural Network approaches use a combination of machine learning algorithms configured as layers of nodes, which process inputs of data or outputs from previous nodes (Pourgholamali, Kahani, Bagheri and Noorian, 2017). A subset of these approaches, Language models are combinations of neural networks trained by predicting blanked-out words in texts (Otter, Medina and Kalita, 2018 ) using a technique called Transformer, which allows for parallel training on multiple processors. Examples of such models include Google's BERT and OpenAI's GPT (Generative Pre-Trained).

The latter, GPT combines transformers with other machine learning models (Vig, 2019). The current iteration, GPT-3 has 175 billion parameters. GPT-3 can recognise grammar, essay structure, and writing genre based on the analysis of very large text datasets. It can be retrained on small datasets to perform tasks such as summarisation and question answering which cannot be done by statistical, unsupervised or supervised learning techniques. GPT-3 can be deployed using natural language prompts that apply the software's rich representations of language on itself to configure its internal neural networks.

In practice, users of GPT can create statements or prompts that describe knowledge tasks. That may include tasks such as "write an academic abstract on Y topic for X journal". This is translated by the program into required software actions that result in text generation, transformation and summarisation to produce a final output. On November 30th 2022, an adapted version of GPT was launched via a simple to use chat interface. ChatGPT, as it is known has been trained using Reinforcement Learning with Human Feedback. ChatGPT has grown to 30 million users in two months, faster than many other digital products (https://www.nytimes.com/2023/02/03/technology/chatgpt-openai-artificial-intelligence.html).

Researchers have begun to use GPT as not merely an analytical tool but a contributor to the academic knowledge creation process as it can assist with core tasks of research such as identifying potential academic contributions, forming and prioritising ideas (Du, Kim, Raheja, Kumar & Kang, 2022). To date, researchers in education (Baidoo-Anu & Owusu Ansah, 2023), medicine (Shen,Heacock, Elias, Hentel, Reig, Shih, & Moy, 2023), and tourism (Carvalho & Ivanov, 2023), among others, are examing the impact of the adoption of LLMs in their domains.

Research, however, has not yet explored the potential for AI to change the underlying practices of academic knowledge creation. The exponential growth of data processing power has led to advances in AI, including self-supervised neural models, that can learn powerful representations from large-scale unstructured data such as text without human supervision. In this manner, they can go beyond analysis by applying these representations to generate outputs in the form of text, audio, images and video (Weisz,Muller, He, & Houde, 2023). For example, in 2019 Springer published the first academic book written by AI (Writer, 2019).

Like other types of digital products, GPT, has catalysed an online community of knowledge and practices (Van de Vrande, De Jong, Vanhaverbeke, & De Rochemont, 2009). This community provides advice on how to utilize software applications in addition to offical support (Cosentino, Izquierdo, and Cabot 2017). Growth in available advice will make core tools more accessible to non-technical individuals, supporting increased adoption even if core technical functionality does not change.

GPT and other language models are poised to be embedded in consumer word processing and other applications (https://blogs.microsoft.com/blog/2023/01/23/microsoftandopenaiextendpartnership/). This can only ensure that the number of users, including academics, will perform knowledge-creation tasks using these tools. Improved language models are under development which are designed to overcome the limitations of existing offerings (https://www.datacamp.com/blog/what-we-know-gpt4). Combined, these drivers suggest that their impact on knowledge creation will continue to grow.

This paper makes an initial contribution based on a scoping review (2018 to January 2023) of how researchers have previously used the language model GPT to assist in the performance of academic knowledge creation tasks beyond the analysis of data. These tasks include writing, editing, reviewing, dataset creation and curation, which have been difficult to perform using earlier ML tools form the basis of creating research outputs, which underpin academic impact, knowledge exchange with industry and educational experiences. Based on a synthesis of these papers, we identify pathways for a future academic research landscape that incorporates wider usage of large language models based on the current modes of adoption in published articles as a CoWriter, Research Assistant and Respondent. The paper concludes with a research and practice agenda for management knowledge creation based on the wider adoption of Large Language models. The paper's focus is on understanding the nature of the current usage of GPT to perform academic tasks. As such, it does not describe the challenges and problems of large language models. It does also not speculate about the extent to which they present machine intelligence or consciousness.

## 3. Method

This research takes the form of a scoping study, a type of literature review that aims to identify sources of evidence in a research area. Unlike systematic reviews, scoping reviews do not focus on a well-defined question and tend to address broader topics in an emerging area (Brogaard, 2021). Scoping studies can provide an initial overview of an area that has not been reviewed before. Given the nature of the approach, the research questions tend to be broad and can include non-peer-reviewed articles, which may limit the robustness of the findings (Pham, Rajić, Greig, Sargeant, Papadopoulos & McEwen, 2014). This scoping review was created to identify research was done that examines the use of GPT as in scholarly knowledge production beyond the analysis of data to support the creation of outputs. We identify academic tasks as belonging to the broad categories identified in Table 1.

Table 1: Categories of Academic Tasks

| Academic Task | Summary |
|---|---|
| Identify area of focus | Scanning and evaluating current academic research |
| Identify possible research directions | Based on the evaluation, identify and prioritise possible research directions |
| Research gap(s) and academic contribution identification | Identify the nature of gaps in existing work (insufficient research, overlooked area, emerging area in need of empirical work or synthesis) and identify potential academic contribution from the planned study. Identify potential outlet for publication. |
| Finalise knowledge development approach | Determine the approach (literature review, secondary or primary data analysis) required to create an academic contribution. Design approach protocol and perform exploratory evaluation. |
| Perform knowledge development activity | Adapt analysis protocol, and document decisions made and findings. Identify academic and industry implications of findings. |
| Create initial research communication artefact | Complete output (paper, presentation), and submit to outlet. |

Adapted from Hope, Downey, Etzioni, Weld & Horvitz (2022).

Records were first identified using the search term GPT, GPT-2 and GPT-3 and academic tasks from table 1 using Scopus, Google Scholar and Semantic Scholar from 2018 to Jan 2023. To limit the search results, the articles had to mention academic tasks, including academic writing, literature summary, text generation, and experiment design. Abstracts were screened by two academics using the platform Rayyan.ai to remove duplicates and remove irrevelant studies. This left 182 studies to be examined in detail for eligibility and were removed if they did not focus on one or more of the categories identified in Table 1.

Figure 1 provides an overview of the knowledge search and identification of the final subset of articles.

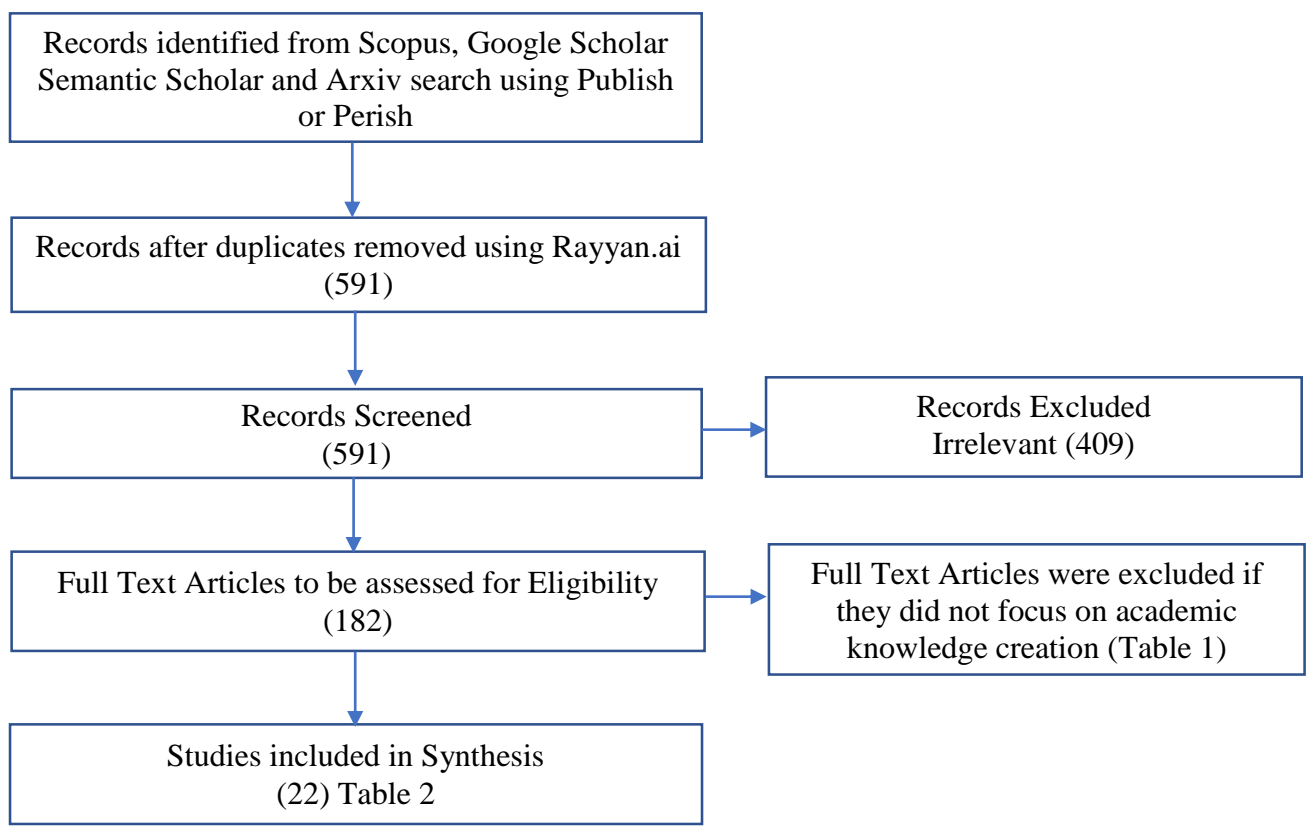

## 4. Findings: GPT in Academic Knowledge production

GPT, like other large language models have a number of impacts on academic knowledge creation. While large language models have been available for some time, previous versions required programming knowledge in order to obtain full benefit of usage. The public availability of GPT via a chat interface has enabled non-programmers to access these advanced tools, thus leading to the democratisation of academic knowledge creation. There is also an exponential increase in available advice on how to create prompts to be able to get the best possible output for a wide variety of tasks. Table 2 summarizes how GPT has been employed for academic knowledge-creation tasks.

Table 2: Papers

| Type | Title and Reference | Classification |
|---|---|---|
| Preprint | Srivastava, M. (2023, January 9). A Day in the Life of ChatGPT as a researcher: Sustainable and Efficient Machine Learning - A Review of Sparsity Techniques and Future Research Directions. https://doi.org/10.31219/osf.io/e9p3g | GPT as Cowriter |
| Preprint | Uchendu, A., Le, T., & Lee, D. (2022). Attribution and Obfuscation of Neural Text Authorship: A Data Mining Perspective. arXiv preprint arXiv:2210.10488. | GPT as Cowriter |
| Journal article | Leippold, M. (2022). Thus spoke GPT-3: Interviewing a large-language model on climate finance. Finance Research Letters, 103617. | GPT as Cowriter |
| Preprint | Liew, A., & Mueller, K. (2022). Using Large Language Models to Generate Engaging Captions for Data Visualizations. arXiv preprint arXiv:2212.14047. | GPT as Cowriter |
| Preprint | Liyanage, V., Buscaldi, D., & Nazarenko, A. (2022). A benchmark corpus for the detection of automatically generated text in academic publications. *arXiv preprint arXiv:2202.02013*. | GPT as Cowriter |
| Journal article | Nowak-Gruca, A. J. (2022). Could an Artificial Intelligence be a Ghostwriter?. Journal of Intellectual Property Rights (JIPR), 27(1), 25-37. | GPT as Cowriter |
| Journal article | Illia, L., Colleoni, E., & Zyglidopoulos, S. (2023). Ethical implications of text generation in the age of artificial intelligence. Business Ethics, the Environment & Responsibility, 32(1), 201-210. | GPT as Cowriter |
| Journal article | Alarie, B., & Cockfield, A. (2021). Will machines replace us?: Machine-authored texts and the future of scholarship. *Law, Technology and Humans*, *3*(2), 5-11. | GPT as Cowriter |
| Conference paper | Tallón-Ballesteros, A. J. (2020). Exploring the potential of GPT-2 for generating fake reviews of research papers. Fuzzy Systems and Data Mining VI: Proceedings of FSDM, 331, 390. | GPT as Cowriter |
| Journal article | Pavlik, J. V. (2023). Collaborating With ChatGPT: Considering the Implications of Generative Artificial Intelligence for Journalism and Media | GPT as Cowriter |

| Type | Title and Reference | Classification |
|---|---|---|
| | Education. *Journalism & Mass Communication Educator*, 10776958221149577. | |
| Conference paper | Sekulić, I., Aliannejadi, M., & Crestani, F. (2022, February). Evaluating mixed-initiative conversational search systems via user simulation. In Proceedings of the Fifteenth ACM International Conference on Web Search and Data Mining (pp. 888-896). | GPT as Respondent |
| Conference paper | Hämäläinen, P., Tavast, M., & Kunnari, A. (2022, March). Neural Language Models as What If?-Engines for HCI Research. In *27th International Conference on Intelligent User Interfaces* (pp. 77-80). | GPT as Respondent |
| Conference paper | Tavast, M., Kunnari, A., & Hämäläinen, P. (2022, March). Language Models Can Generate Human-Like Self-Reports of Emotion. In 27th International Conference on Intelligent User Interfaces (pp. 69-72). | GPT as Respondent |
| Conference paper | Meyer, S., Elsweiler, D., Ludwig, B., Fernandez-Pichel, M., & Losada, D. E. (2022, July). Do We Still Need Human Assessors? Prompt-Based GPT-3 User Simulation in Conversational AI. In Proceedings of the 4th Conference on Conversational User Interfaces (pp. 1-6). | GPT as Respondent |
| Journal article | Salehi, P., Hassan, S. Z., Lammerse, M., Sabet, S. S., Riiser, I., Røed, R. K., ... & Riegler, M. A. (2022). Synthesising a talking child avatar to train interviewers working with maltreated children. Big Data and Cognitive Computing, 6(2), 62. | GPT as Respondent |
| Preprint | Horton, J. J. (2023). Large Language Models as Simulated Economic Agents: What Can We Learn from Homo Silicus?. arXiv preprint arXiv:2301.07543. | GPT as Respondent |
| Journal article | Jaimovitch-López, G., Ferri, C., Hernández-Orallo, J., Martínez-Plumed, F., & Ramírez-Quintana, M. J. (2022). Can language models automate data wrangling?. *Machine Learning*, 1-30. | GPT as Research Assistant |
| Journal article | Hernandez, I., & Nie, W. (2022). The AI-IP: Minimising the guesswork of personality scale item development through artificial intelligence. *Personnel Psychology*. | GPT as Research Assistant |
| Journal article | Lee, P., Fyffe, S., Son, M., Jia, Z., & Yao, Z. (2022). A Paradigm Shift from "Human Writing" to "Machine Generation" in Personality Test Development: an Application of State-of-the-Art Natural Language Processing. Journal of Business and Psychology, 1-28. | GPT as Research Assistant |
| Preprint | Ye, J., Gao, J., Li, Q., Xu, H., Feng, J., Wu, Z., ... & Kong, L. (2022). Zerogen: Efficient zero-shot | GPT as Research Assistant |

| Type | Title and Reference | Classification |
|---|---|---|
|  | learning via dataset generation. arXiv preprint arXiv:2202.07922. |  |
| Conference paper | Bellan, P., Dragoni, M., & Ghidini, C. (2022). Experiment Maker: a Tool to create Experiments with GPT-3 easily. EKAW'22: Companion Proceedings of the 23rd International Conference on Knowledge Engineering and Knowledge Management, September 26–29, 2022, Bozen-Bolzano, IT | GPT as Research Assistant |
| Journal article | Kansteiner, W. (2022). Digital doping for historians: can history, memory, and historical theory be rendered artificially intelligent? History and Theory.https://doi.org/10.1111/hith.12282 | GPT as Research Assistant |

Three themes emerged from the literature:
1) GPT as a *Co-Writer* in which the tool was deployed to complete academic outputs (paper, presentation) to scan and evaluate current academic research, identify and prioritise possible research directions.
2) GPT as a *Research Assistant* determines the approach (literature review, secondary or primary data analysis) required to create academic contribution; designs approach protocol and performs exploratory evaluation; performs evaluation or analysis; documents decisions made and findings; identifies academic and industry implications of findings.
3) GPT as a *Respondent* in which GPT was used as a source of simulated respondents and systems.

### 4.1 GPT as a Co-Writer

Previous Natural Language Generation (NLG) software systems followed rule-based systems. Large Language models learn representations from large text collections, in the case of GPT, 175 billion parameters. The use of these tools can enable researchers to synthesise related work as well as expand the exploration of problem spaces to adjacent and parallel fields (Alarie and Cockfield, 2021). Many management phenomena are examined differently in different fields and even related fields. The use of GPT (Table 2) to summarise and synthesise knowledge can enable research teams at the initial stage to deploy arguments based on developments in parallel domains in business and management or broadly across the social sciences.

Researchers have speculated on the role of GPT as a disguised ghost-writer in academic research (Srivastava, 2022). Although some software applications already exist (e.g. https://writer.com/ai-content-detector/), text created by AI is difficult to detect, and formal response requires the coordination of multiple technological and social institutions in order to establish acceptable use of text generation (Illia, Colleoni and Zyglidopoulos, 2022). As these models become bigger, the act of Authorship Attribution (human or machine becomes more difficult (Nowak-Gruca, 2022). The introduction of ChatGPT forced many academic publishers to adopt formal policies towards AI-generated text and AI authorship of academic publications. The consensus is that GPT or other LLM cannot be a co-author and must be treated as a tool. Moreover, the use of AI to generate texts needs to be explicitly acknowledged but the human authors take full responsibility for the manuscript's content.

In addition to core academic texts, GPT has been used to create academic peer review reports (Bartoli and Medvet, 2020). In both cases, the team adapted the model to academic domains by training them on domain-specific text so that they can create outputs in the required format and style. The rate of knowledge production is increasing and in many domains, the number of papers that are available on a weekly or monthly basis frequently exceeds the capacity of individual researchers to meaningfully absorb. There are also potential modalities of fraud based on the creation of fake review reports (Uchendu, Le and Lee, 2022). Since reviewers are anonymous, review reports for papers can be generated without attribution, that may be used to reduce trust in the academic process. These tools also have indirect impacts. For researchers who depend on public text sources to generate outputs, such as social media these tools may pose a problem as they will be an increasing amount of fake reviews and posts on online platforms (Karanjai, 2022). In both cases we will have knowledge being created by tools that do not have a mind, worldview or perspective and are simply presenting words based on statistical inference rather than on meaning.

Far more dangerous are the use of these tools to create false or misleading information from external actors, which may result increased amount of academic hoaxes (Al-Khatib & Teixeira da Silva, 2016). These frauds have been popular where politically motivated actors create fake articles in an attempt to show academic biases and flaws (Piedra, 2019). The ability to generate plausible text in the style of targeted journals as well as plausible data sets will increase the volume of these hoaxes and academia will have to create enhanced ways of identifying these hoaxes.

**4.2 GPT as a research assistant**

This stream of research (Table 2) identifies the potential for these tools to support literature search, data preparation, transformation and synthesis tasks performed by academics. For synthesis, these tools can create summaries of existing text, including examination of arguments (Alarie & Cockfield, 2021; Illia, Colleoni, & Zyglidopoulos, 2023). Data quality improvement by actions such as cleaning and curation are critical for computational analyses of large datasets. Attempts have been made to utilise machine learning approaches to automate this process, but they face limitations of determining relevance and can result in errors. Due to the complex nature, a significant amount of this work is done via crowdsourcing or hiring of data cleaning staff. Academics have used GPT to perform tasks that require domain knowledge on unstructured data, including cleaning, formatting and exploratory analysis (Jaimovitch-López, Ferri, Hernández-Orallo, Martínez-Plumed & Ramírez-Quintana, 2022). GPT has been able to identify required elements in text, fill in data that is missing using a semantic approach, learn transformation functions and identify anomalies in multimodal data sets given a few examples from researchers. In this way, they automate data preprocessing tasks in a manner that can shape subsequent research by reducing the time to apply multiple transformation approaches which can support a greater range of analytical tasks (De Bie et al, 2022).

In addition to tasks on unstructured data, GPT has been used to support conventional quantitative and qualitative analyses. For the former, GPT has been used to create items for scale development by generating a large number of options which were then refined using stated researcher priorities (Hernandez & Nie, 2022). Scale development is a complex research task that can be limited by the team's capacity for identifying potential items, creating appropriate descriptions, ensuring validity and identifying correlations among items. GPT outputs were found to be equivalent to those created using traditional approaches (Hernandez & Nie, 2022).

GPT has also been used as an approach to explore research options. The tool has been used to examine potential respondent behaviour and interactions by querying it's internal text representations. In contrast to traditional data collection approaches, GPT was able to support the development of qualitative and quantitative data collection as well as enable researchers in order to explore confactual what if questions (Hämäläinen, Tavast, & Kunnari, 2022). In this way, GPT was used to improve research methodology, not simply perform analyses. Even further, in ZeroGen, GPT was used to create a subset of itself, called a tiny task model (TAM) to perform specific types of analyses when an existing machine learning model did not exist (Ye, Gao, Li, Xu, Feng, Wu, & Kong, 2022). Unlike previous approaches, the model did not require external training data in order to create a tool that could analyse specific types of data. In this area, GPT has also been used recursively on itself to design large language experiments (Bellan, Dragoni, & Ghidini, 2022).

### 4.3 GPT as Respondent

GPT has itself become a respondent to create papers. Without any additional data sources, researchers have queried GPT on perspective on issues such as climate change to identify biases or dominant perspectives in its internal language model (Leippold, 2022). GPT has also acted as a respondent to interview questions in traditional academic research (Iskender, 2023).

GPT has also been acted as a participant in experiments, replacing crowdsourced workers in computing and economics research (Bellan, Dragoni, and Ghidini, 2022). In the former, GPT has acted as a surrogate user in a conversational search system to ask questions of a given information system and evaluate the usefulness of the answers (Meyer, Elsweiler, Ludwig, Fernandez-Pichel & Losada, 2022). In the latter, GPT has been used to replicate findings from famous economic experiments (Horton, 2023).

While GPT does not have a worldview, it may exhibit emergent behaviour based on its training data. These idiosyncrasies are not seen as a limitation but as a benefit for researchers seeking to explore complex behaviours. In this way, GPT can provide simulated responses to questions of emotions that a rich descriptions but are entirely synthetic. These responses can be used to enrich existing datasets or to provide a basis for comparison to extend theoretical work (Ye et al., 2022). This is of particular value where few respondents are available or respondents may be unresponsive. In this mode, GPT has been used to create simulated responders for sensitive topics that allow researchers to explore these areas without causing harm (Salehi et al., 2022). These respondents need not be individuals as group interactions can also be modelled (Hamilton 2023). However, this means that the studies that are based on simulated responses and emotions do not evaluate actual human perceptions and emotions; hence, the validity of the findings of such papers will be limited to the AI domain and they should not be generalised to humans. A related stream of research identifies the biases of Chat GPT as a respondent on specific subjects. The tool has been used to answer Positive and Negative Affect Schedule (PANAS) items (Lee, Fyffe, Son, Jia and Yao, 2022).

### 5.0 Discussion and Research Agenda
Large language models like GPT create new capacities and constraints to Business and Management academics involved in research output creation. The above themes suggest that LLMs like GPT can increase the capacity of academic teams to perform research. Management academic researchers are increasingly required to provide knowledge that is not just rigorous but impactful (Wickert et al., 2021). Management research has been criticised for having a gap between researchers and practitioners, as managers rarely read articles in top management journals due to their theoretical nature with limited practical value (Kieser, Nicolai, & Seidl, 2015). Additionally, the focus is on publishing in highly ranked/high-impact journals to increase institutional status with financial and reputational benefits. By increasing the capacity of researchers to deliver research, Large Language models may enable the field to enact its societal responsibilities by being able to generate rigorous impactful research. This benefit may be tempered by the increasing volume of outputs from researchers using simulated outputs in order to meet institutional status requirements.

## 5.1 Large Language Models and Academic Capacity Expansion

The use of these tools to summarise and synthesise knowledge can enable research teams at the initial stage to gain some insight into methodological and theoretical developments in parallel domains in business and management or broadly across the social sciences. By expanding the exploration of the problem space, academic contributions could be based on a broader conceptual base range. In this way academic silos can be broken down by conceptual frameworks enriched by contributions from parallel fields (Pavlik, 2023). In areas where there may a limited number of respondents, such as niche populations or difficult populations, these tools can be used to help refine data collection instruments or to generate simulated data (Salehi, Hassan, Lammerse, Sabet, Riiser, Røed & Riegler, 2022). However, in the, the validity of such studies might be questionable and new modes of verification in addition to conceptual validity and triangulation must be created to examine the validity of computer generated responses.

Large language models provide the opportunity to create new types of outputs based on syntesis of extant research. These technologies can be applied as as a precursor to or a supplement to a systematic literature review. The rate of scholarly production is ever increasing (World Bank, n.d.) and academics may find it difficult to keep up with the body of knowledge (Johann, Raabe, & Rauhut, 2022). Furthermore, articulating a distinct academic contribution may require academics to summarise and synthesise different bodies of knowledge, which can be difficult (Lindgreen, Di Benedetto, Clarke, Evald, Bjørn-Andersen & Lambert, 2021). The use of these language models as initial summarisation tools in their research assistant role may be of value. Instead of narrowing down to a small number of articles for synthesis, researchers can explore broader questions based on contributions that are embedded in different types of knowledge. This can create a new type of systematic integrated review that extends the current manual approach using the summarization capabilities of these tools (Elsbach & van Knippenberg, 2020).

The second new type of output may be based on prompt programming. Researchers have published articles that combine text, code and data (Hildebrand, Efthymiou, Busquet, Hampton, Hoffman, & Novak, 2020). The way in which GPT is accessed is via prompting which may be a series of instructions that can be increasingly refined to provide feedback to the model (Zhou, Muresanu, Han, Paster, Pitis, Chan & Ba, 2022). Future papers may include a structured description of prompts and responses along with a recording of text generation in real time to enable replication of research.

Future academic outputs may be based on entirely new methodologies facilitated by GPT. Netnography, for example, adapted the idea of ethnography to online communities and interactions (Kozinets, 2020). Language models may be used in a similar manner to query themselves in a new form of ethnography. In this case, the "respondent" is an aggregated body of statistical patterns derived from online text and the researcher is querying statistical patterns in the text to gain some insight into what common knowledge or widely held perspectives are on a given topic or area. When prompted by the authors, ChatGPT suggested such netnographic approach to be named "AI-based netnography" ot "AI-driven netnography".

GPT can also be prompted to act in different rules and can be used to extend existing analyses that may require the creation of narratives from a distinct population that is difficult to access (Salehi et al., 2022). In this role, they may also be a distinct form of social simulation. Existing simulation models use agent-based modelling or system dynamics to model associations among numerical variables. A large language model does not need such an abstraction and can directly simulate interactions among simulated characters that it creates (Hamilton, 2023). This may be a new way of performing social simulation activities that is not based on simplification of a problem, but on simulating scenarios based on a richer form of qualitative description.

## 5.2 Large Language Models and Academic Capacity Constraints

The use of these tools can increase the power of technological companies over academic knowledge production. The development of large language models is funded by very large commercial organisations and are not public goods (Bender, Gebru, McMillan-Major & Shmitchell, 2021). The information provided to these tools via researcher usage help train the systems to improve knowledge creation (Pavlik, 2023). Increasing the use of these tools in academic publications allows them to become better at creating academic writing which ironically will increase their power over knowledge production. For researchers who rely on technological companies for data access, changes in the terms of service can force cancellation of planned initiatives, in drastic cases affecting an entire subdomain of research (Bruns, 2019). Further, as large language models are inscrutable, they may embed training dataset biases on outputs, indirectly shaping academic research (Horton, 2023).

For researchers who depend on public text sources to generate outputs such as social media these tools may pose a problem as they will be an increasing amount of fake reviews and posts on online platforms. The problem of fake data will be a major one since academics, students and public can create plausible sounding data using simple prompts for interviews or take existing data sets and modify them quite simply to create a plausible seeming data set that is then analysed and presented as if it were collected from real individuals (Tallón-Ballesteros, 2020). The increasing reliance of academics on platforms such as Mechanical Turk is another potential concern since these respondents may also use large language models in order to generate responses to be paid for surveys or experimental participation (Tavast, Kunnari, & Hämäläinen, 2022).

## 6. Concluding remarks

Large Language Models are taking academia like a storm. While there are many fears about them, there are significant benefits in terms of knowledge creation. New research methodology, increased output, better quality of the research output, and new insights and only some of the potential impacts of these technologies on academic knowledge creation. One thing to remember is that LLMs are nothing else but tools, sophisticated but yet tools. They do not have consciousness and cannot take responsibility for the written text. Therefore, they cannot be listed as co-authors of academic publications. However, they can assist in all stages of the research process, making it more effective and efficient. Hence, in the future, researchers may not go the way of horses (Brynjolffson & McAfee, 2015) but rather researchers that utilise AI (in this case LLMs) will outperform researchers that do not on traditional metrics. Thus, LLMs may be a source of competitive advantage in academia and skills to use LLMs will be part of researchers' near future core competences. Research methods modules at universities will need to incorporate LLM-based research methodologies and skills in order to equip the future researchers with the necessary research skills. As the LLMs develop, so will the researchers in a co-evolution game that has no end or winner but fuzzy rules, evolution path and knowledge co-creation.

**References:**

Alarie, B., & Cockfield, A. (2021). Will machines replace us?: Machine-authored texts and the future of scholarship. Law, Technology and Humans, 3(2), 5-11.

Al-Khatib, A., & Teixeira da Silva, J. A. (2016). Stings, hoaxes and irony breach the trust inherent in scientific publishing. *Publishing Research Quarterly*, *32*(3), 208-219.

Baidoo-Anu, D., & Owusu Ansah, L. (2023). Education in the Era of Generative Artificial Intelligence (AI): Understanding the Potential Benefits of ChatGPT in Promoting Teaching and Learning. *Available at SSRN 4337484*.

Bellan, P., Dragoni, M., & Ghidini, C. (2022). Experiment Maker: a Tool to create Experiments with GPT-3 easily. EKAW'22: Companion Proceedings of the 23rd International Conference on Knowledge Engineering and Knowledge Management, September 26–29, 2022, Bozen-Bolzano, IT

Bender, E. M., Gebru, T., McMillan-Major, A., & Shmitchell, S. (2021, March). On the Dangers of Stochastic Parrots: Can Language Models Be Too Big?. In Proceedings of the 2021 ACM Conference on Fairness, Accountability, and Transparency (pp. 610-623).

Block, J. H., Fisch, C., Kanwal, N., Lorenzen, S., & Schulze, A. (2022). Replication studies in top management journals: An empirical investigation of prevalence, types, outcomes, and impact. *Management Review Quarterly*, 1-26.

Brogaard, L. (2021). Innovative outcomes in public-private innovation partnerships: a systematic review of empirical evidence and current challenges. *Public Management Review*, *23*(1), 135-157.

Bruns, A. (2019). After the 'APIcalypse': Social media platforms and their fight against critical scholarly research. *Information, Communication & Society*, *22*(11), 1544-1566.

Brynjolffson, E., & McAfee, A. (2015). Will Humans Go the Way of Horses? Labor in the Second Machine Age. *Foreign Affairs*, *94*(4), 8-14.

Carvalho, I., & Ivanov, S. (2023). ChatGPT for tourism: applications, benefits, and risks. *Tourism Review*, https://doi.org/10.1108/TR-02-2023-0088 (in press),

Cosentino, V., Izquierdo, J.L.C. and Cabot, J., 2017. A systematic mapping study of software development with GitHub. IEEE Access, 5, pp.7173-7192.

Du, W., Kim, Z. M., Raheja, V., Kumar, D., & Kang, D. (2022). Read, Revise, Repeat: A System Demonstration for Human-in-the-loop Iterative Text Revision. arXiv preprint arXiv:2204.03685.

Elsbach, K. D., & van Knippenberg, D. (2020). Creating high-impact literature reviews: An argument for 'integrative reviews'. *Journal of Management Studies*, *57*(6), 1277-1289.

Ghahramani, Z. (2015). Probabilistic machine learning and artificial intelligence. *Nature*, 521(7553), p.452.

Grace, K., Salvatier, J., Dafoe, A., Zhang, B. and Evans, O. (2017). When will AI exceed human performance? Evidence from AI experts. arXiv preprint arXiv:1705.08807.

Hämäläinen, P., Tavast, M., & Kunnari, A. (2022, March). Neural Language Models as What If?-Engines for HCI Research. In 27th International Conference on Intelligent User Interfaces (pp. 77-80).

Hamilton, S. (2023). Blind Judgement: Agent-Based Supreme Court Modelling With GPT. *arXiv preprint arXiv:2301.05327*.

Hernandez, I., & Nie, W. (2022). The AI-IP: Minimising the guesswork of personality scale item development through artificial intelligence. *Personnel Psychology*. DOI: 10.1111/peps.12543

Hernández-Orallo, J. (2019). Gazing into Clever Hans machines. *Nature Machine Intelligence*, 1, 172-173.

Hope, T., Downey, D., Etzioni, O., Weld, D. S., & Horvitz, E. (2022). A Computational Inflection for Scientific Discovery. *arXiv preprint arXiv:2205.02007*.

Horton, J. J. (2023). Large Language Models as Simulated Economic Agents: What Can We Learn from Homo Silicus?. arXiv preprint arXiv:2301.07543.

Illia, L., Colleoni, E., & Zyglidopoulos, S. (2023). Ethical implications of text generation in the age of artificial intelligence. Business Ethics, the Environment & Responsibility, 32(1), 201-210.

Iskender, A. (2023). Holy or Unholy? Interview with Open AI's ChatGPT. *European Journal of Tourism Research*, 34, Article number 3414. https://doi.org/10.54055/ejtr.v34i.3169

Jaimovitch-López, G., Ferri, C., Hernández-Orallo, J., Martínez-Plumed, F., & Ramírez-Quintana, M. J. (2022). Can language models automate data wrangling?. Machine Learning, 1-30.

Johann, D., Raabe, I. J., & Rauhut, H. (2022). Under pressure: The extent and distribution of perceived pressure among scientists in Germany, Austria, and Switzerland. *Research Evaluation*, *31*(3), 385-409.

Karanjai, R. (2022). Targeted Phishing Campaigns using Large Scale Language Models. arXiv, available at: https://doi.org/10.48550/arxiv.2301.00665 (accessed 3 February 2023)

Kansteiner, W. (2022). Digital doping for historians: can history, memory, and historical theory be rendered artificially intelligent? History and Theory.https://doi.org/10.1111/hith.12282

Kieser, A., Nicolai, A., & Seidl, D. (2015). The practical relevance of management research: Turning the debate on relevance into a rigorous scientific research program. *Academy of Management annals*, *9*(1), 143-233.

Kozinets, R. (2020). Netnography: The essential guide to qualitative social media research. Sage.

Leavitt, K., Schabram, K., Hariharan, P., & Barnes, C. M. (2021). Ghost in the machine: On organisational theory in the age of machine learning. *Academy of Management Review*, *46*(4), 750-777.

Lee, P., Fyffe, S., Son, M., Jia, Z., & Yao, Z. (2022). A Paradigm Shift from "Human Writing" to "Machine Generation" in Personality Test Development: an Application of State-of-the-Art Natural Language Processing. Journal of Business and Psychology, 1-28.

Leippold, M. (2022). Thus spoke GPT-3: Interviewing a large-language model on climate finance. Finance Research Letters, 103617.

Liew, A., & Mueller, K. (2022). Using Large Language Models to Generate Engaging Captions for Data Visualizations. arXiv preprint arXiv:2212.14047.

Lindgreen, A., Di Benedetto, C. A., Clarke, A. H., Evald, M. R., Bjørn-Andersen, N., & Lambert, D. M. (2021). How to define, identify, and measure societal value. Industrial Marketing Management, 97, A1-A13.

Liyanage, V., Buscaldi, D., & Nazarenko, A. (2022). A benchmark corpus for the detection of automatically generated text in academic publications. arXiv preprint arXiv:2202.02013.

Martinez-Torres, M. D. R., & Toral, S. L. (2019). A machine learning approach for the identification of the deceptive reviews in the hospitality sector using unique attributes and sentiment orientation. *Tourism Management*, *75*, 393-403.

Meyer, C., Cohen, D., & Nair, S. (2020). From automats to algorithms: the automation of services using artificial intelligence. Journal of Service Management, 31(2), 145-161.

Mohammad, A. S., Jaradat, Z., Mahmoud, A. A., & Jararweh, Y. (2017). Paraphrase identification and semantic text similarity analysis in Arabic news tweets using lexical, syntactic, and semantic features. *Information Processing & Management*, *53*(3), 640-652.

Müller, V. C., & Bostrom, N. (2016). Future progress in artificial intelligence: A survey of expert opinion. *Fundamental issues of artificial intelligence*, 555-572.

Nowak-Gruca, A. J. (2022). Could an Artificial Intelligence be a Ghostwriter?. Journal of Intellectual Property Rights (JIPR), 27(1), 25-37.

Otter, D. W., Medina, J. R., & Kalita, J. K. (2020). A survey of the usages of deep learning for natural language processing. *IEEE transactions on neural networks and learning systems*, *32*(2), 604-624.

Pavlik, J. V. (2023). Collaborating With ChatGPT: Considering the Implications of Generative Artificial Intelligence for Journalism and Media Education. Journalism & Mass Communication Educator, 10776958221149577.

Pham, M. T., Rajić, A., Greig, J. D., Sargeant, J. M., Papadopoulos, A., & McEwen, S. A. (2014). A scoping review of scoping reviews: advancing the approach and enhancing the consistency. *Research synthesis methods*, *5*(4), 371-385.

Piedra, L. M. (2019). The gift of a hoax. *Qualitative Social Work*, 18(2), 152-158.

Pourgholamali, F., Kahani, M., Bagheri, E. and Noorian, Z., 2017. Embedding unstructured side information in product recommendation. Electronic Commerce Research and Applications, 25, pp.70-85.

Rodríguez, J., Semanjski, I., Gautama, S., Van de Weghe, N., & Ochoa, D. (2018). Unsupervised hierarchical clustering approach for tourism market segmentation based on crowdsourced mobile phone data. *Sensors*, *18*(9), 2972.

Russell, S. J. (2010). *Artificial intelligence a modern approach*. Pearson Education, Inc..

Sekulić, I., Aliannejadi, M., & Crestani, F. (2022, February). Evaluating mixed-initiative conversational search systems via user simulation. In Proceedings of the Fifteenth ACM International Conference on Web Search and Data Mining (pp. 888-896).

Salehi, P., Hassan, S. Z., Lammerse, M., Sabet, S. S., Riiser, I., Røed, R. K., ... & Riegler, M. A. (2022). Synthesising a talking child avatar to train interviewers working with maltreated children. Big Data and Cognitive Computing, 6(2), 62.

Shen, Y., Heacock, L., Elias, J., Hentel, K. D., Reig, B., Shih, G., & Moy, L. (2023). ChatGPT and Other Large Language Models Are Double-edged Swords. *Radiology*, 230163.

Srivastava, M. (2023, January 9). A Day in the Life of ChatGPT as a researcher: Sustainable and Efficient Machine Learning - A Review of Sparsity Techniques and Future Research Directions. https://doi.org/10.31219/osf.io/e9p3g

Tallón-Ballesteros, A. J. (2020). Exploring the potential of GPT-2 for generating fake reviews of research papers. Fuzzy Systems and Data Mining VI: Proceedings of FSDM, 331, 390.

Tavast, M., Kunnari, A., & Hämäläinen, P. (2022, March). Language Models Can Generate Human-Like Self-Reports of Emotion. In 27th International Conference on Intelligent User Interfaces (pp. 69-72).

Uchendu, A., Le, T., & Lee, D. (2022). Attribution and Obfuscation of Neural Text Authorship: A Data Mining Perspective. arXiv preprint arXiv:2210.10488.

Vig, J. (2019). A multiscale visualization of attention in the transformer model. *arXiv preprint arXiv:1906.05714*.

Van de Vrande, V., De Jong, J. P., Vanhaverbeke, W., & De Rochemont, M. (2009). Open innovation in SMEs: Trends, motives and management challenges. *Technovation*, *29*(6-7), 423-437.

Weisz, J. D., Muller, M., He, J., & Houde, S. (2023). Toward General Design Principles for Generative AI Applications. arXiv preprint arXiv:2301.05578.

Wickert, C., Post, C., Doh, J. P., Prescott, J. E., & Prencipe, A. (2021). Management research that makes a difference: Broadening the meaning of impact. *Journal of Management Studies*, *58*(2), 297-320.

World Bank (n.d.) Scientific and technical journal articles 2000-2018. Retrieved from https://data.worldbank.org/indicator/IP.JRN.ARTC.SC

Writer, B. (2019). *Lithium-Ion Batteries. A Machine-Generated Summary of Current Research*. Cham: Springer. Retrieved from https://link.springer.com/content/pdf/10.1007%2F978-3-030-16800-1.pdf

Yadav, M. S. (2018). Making emerging phenomena a research priority. *Journal of the Academy of Marketing Science*, *46*, 361-365.

Ye, J., Gao, J., Li, Q., Xu, H., Feng, J., Wu, Z., ... & Kong, L. (2022). Zerogen: Efficient zero-shot learning via dataset generation. arXiv preprint arXiv:2202.07922.

Zhang, K., Chen, Y. and Li, C., 2019. Discovering the tourists' behaviors and perceptions in a tourism destination by analysing photos' visual content with a computer deep learning model: The case of Beijing. *Tourism Management*, 75, pp. 595-608.

Zhou, Y., Muresanu, A. I., Han, Z., Paster, K., Pitis, S., Chan, H., & Ba, J. (2022). Large language models are human-level prompt engineers. *arXiv preprint arXiv:2211.01910*.